\documentclass[pdflatex,sn-mathphys-num]{sn-jnl}
\usepackage{graphicx}
\usepackage{multirow}
\usepackage{amsmath,amssymb,amsfonts}
\usepackage{amsthm}
\usepackage{mathrsfs}
\usepackage[title]{appendix}
\usepackage{xcolor}
\usepackage{textcomp}
\usepackage{manyfoot}
\usepackage{booktabs}
\usepackage{algorithm}
\usepackage{algorithmicx}
\usepackage{algpseudocode}
\usepackage{listings}
\usepackage{graphicx}
\usepackage{multirow}
\usepackage{amsmath,amssymb,amsfonts}
\usepackage{amsthm}
\usepackage{mathrsfs}
\usepackage[title]{appendix}
\usepackage{xcolor}
\usepackage{textcomp}
\usepackage{manyfoot}
\usepackage{booktabs}
\usepackage{algorithm}
\usepackage{hyperref}
\usepackage[figuresright]{rotating}
\usepackage{algorithmicx}
\usepackage{algpseudocode}
\usepackage{listings}
\usepackage{cleveref}
\usepackage{ulem}
\usepackage{xspace}
\usepackage[version=4]{mhchem}
\usepackage{subcaption}
\usepackage{multirow}
\usepackage{array}
\usepackage{float}


\usepackage{alphalph}
\usepackage{fnpct}

\Crefname{table}{\text{Table}}{\text{Tables}}
\Crefname{figure}{\text{Fig.}}{\text{Figs.}}
\Crefname{section}{\text{Section}}{\text{Sections}}
\Crefname{equation}{\text{Eq.}}{\text{Eqs.}}

\begin{document}

\title[Article Title]{A Foundational Generative Model for Breast Ultrasound Image Analysis}

\author[1]{\fnm{Haojun} \sur{Yu}}\nomail\equalcont{These authors contributed equally to this work.}
\author[1]{\fnm{Youcheng} \sur{Li}}\nomail\equalcont{These authors contributed equally to this work.}
\author[2]{\fnm{Nan} \sur{Zhang}}\nomail\equalcont{These authors contributed equally to this work.}
\author[3]{\fnm{Zihan} \sur{Niu}}\nomail\equalcont{These authors contributed equally to this work.}
\author[4]{\fnm{Xuantong} \sur{Gong}}\nomail
\author[3]{\fnm{Yanwen} \sur{Luo}}\nomail
\author[5]{\fnm{Haotian} \sur{Ye}}\nomail
\author[5]{\fnm{Siyu} \sur{He}}\nomail
\author[1]{\fnm{Quanlin} \sur{Wu}}\nomail
\author[2]{\fnm{Wangyan} \sur{Qin}}\nomail
\author[3]{\fnm{Mengyuan} \sur{Zhou}}\nomail
\author[4]{\fnm{Jie} \sur{Han}}\nomail
\author[3]{\fnm{Jia} \sur{Tao}}\nomail
\author[8]{\fnm{Ziwei} \sur{Zhao}}\nomail
\author[1]{\fnm{Di} \sur{Dai}}\nomail
\author*[1]{\fnm{Di} \sur{He}}\email{dihe@pku.edu.cn}
\author*[8]{\fnm{Dong} \sur{Wang}}\email{dong.wang@yizhun-ai.com}
\author*[7]{\fnm{Binghui} \sur{Tang}}\email{tbh691203@163.com}
\author*[2]{\fnm{Ling} \sur{Huo}}\email{hlbcus@163.com}
\author*[5]{\fnm{James} \sur{Zou}}\email{jamesz@stanford.edu}
\author*[3]{\fnm{Qingli} \sur{Zhu}}\email{zhuqingli@pumch.cn}
\author*[4,6]{\fnm{Yong} \sur{Wang}}\email{wangyong@cicams.ac.cn}
\author*[1]{\fnm{Liwei} \sur{Wang}}\email{wanglw@pku.edu.cn}

\affil[1]{\orgname{Peking University}}
\affil[2]{\orgname{Peking University Cancer Hospital \& Institute}}
\affil[3]{\orgname{Peking Union Medical College Hospital}}
\affil[4]{\orgname{Cancer Hospital, Chinese Academy of Medical Sciences}}
\affil[5]{\orgname{Stanford University}}
\affil[6]{\orgname{The First Affliated Hospital of  China Medical University}}
\affil[7]{\orgname{Nanchang People's Hospital}}
\affil[8]{\orgname{Yizhun Medical AI Co., Ltd}}

\abstract{
    Foundational models have emerged as powerful tools for addressing various tasks in clinical settings. However, their potential development to breast ultrasound analysis remains untapped. 
    In this paper, we present BUSGen, the first foundational generative model specifically designed for breast ultrasound image analysis. Pretrained on over 3.5 million breast ultrasound images, BUSGen has acquired extensive knowledge of breast structures, pathological features, and clinical variations.
    With few-shot adaptation, BUSGen can generate repositories of realistic and informative task-specific data, facilitating the development of models for a wide range of downstream tasks. 
    Extensive experiments highlight BUSGen's exceptional adaptability, significantly exceeding real-data-trained foundational models in breast cancer screening, diagnosis, and prognosis.
    In breast cancer early diagnosis, our approach outperformed all board-certified radiologists (n=9), achieving an average sensitivity improvement of 16.5\% (P-value$<$0.0001).
    Additionally, we characterized the scaling effect of using generated data which was as effective as the collected real-world data for training diagnostic models. Moreover, extensive experiments demonstrated that our approach improved the generalization ability of downstream models. Importantly, BUSGen protected patient privacy by enabling fully de-identified data sharing, making progress forward in secure medical data utilization. An online demo of BUSGen is available at \href{https://aibus.bio/}{https://aibus.bio}.
}

\maketitle

\section{Main}
\label{sec:main}

Breast cancer is one of the most prevalent cancers, and remains a significant health threat for females worldwide~\cite{siegel2023cancer,chhikara2023global,xia2022cancer}.
Ultrasound is an essential imaging modality to reduce breast cancer-related mortality, which is widely adopted~\cite{sickles2013acr,ohuchi2016sensitivity} for its non-invasiveness, convenience, and higher sensitivity compared to mammography, especially in patients with dense breasts or younger ages ~\cite{berg2012detection,shen2015multi,brem2015screening,sood2019ultrasound,park2020pan}. 
However, the accurate interpretation of breast ultrasound images is challenging and time-consuming due to the complexity of biological structures, the subtlety of pathological semantics, and variability among clinical scenarios~\cite{weese2016four,gonzalez2016review,geirhos2020shortcut,cao2023large,goetz2024generalization}. Even experienced clinicians could make mistakes in interpreting the images, potentially delaying early diagnosis, hindering timely intervene to tumor progression, and compromising treatment plans.

Recent breakthroughs in deep learning have achieved success in medical domains, such as analysis of chest X-ray images~\cite{tiu2022expert,huang2023visual}, pathological images~\cite{xu2024whole}, and retinal images~\cite{de2018clinically,zhou2023foundation}, based on large scale high-quality datasets. However, breast ultrasound analysis remains under-explored by deep learning models, as publicly available breast ultrasound images are limited~\cite{yap2017automated,al2020dataset,lin2022new} due to the privacy and legal constraints~\cite{malin2013biomedical,shen2021artificial}. 
Therefore, the deep learning models for breast ultrasound analysis face challenges such as over-fitting~\cite{wu2019machine,ozaki2022deep,zuluaga2021cnn}, lack of generalization ability~\cite{ong2024shortcut}, learning inefficiency of rare conditions~\cite{shen2021artificial}, and lack of data transparency~\cite{shen2021artificial,namli2024scalable}.
Meanwhile, deep learning is undergoing a paradigm shift from task-specific schemas to foundational models~\cite{brown2020language,ouyang2022training,rombach2022high,team2023gemini,bommasani2021opportunities,zhou2023foundation,xu2024whole,bluethgen2024vision,wang2024self}, which leverage knowledge transfer from extensive and diverse pretraining data to support broad downstream tasks. Consequently, a foundational generative model is essential for enabling precision breast ultrasound analysis.

In this paper, we introduce BUSGen (\textbf{B}reast \textbf{U}ltra\textbf{S}ound \textbf{Gen}erative model), the first foundational generative model designed for breast ultrasound image analysis, which significantly improves a wide range of essential tasks associated with breast cancer screening~\cite{mckinney2020international,lee2010breast,marmot2013benefits}, diagnosis~\cite{qian2021prospective,pacile2020improving,dembrower2020effect}, and prognosis~\cite{zhou2020lymph,tafreshi2010molecular,zhao2020molecular,tsang2020molecular} (\Cref{fig:framework}).
The generative capabilities of BUSGen are rendered by the component of a powerful diffusion model~\cite{ho2020denoising}, pretrained on an extensive dataset encompassing diverse sources, regions, and subpopulations. As shown in \Cref{fig:framework}a, this dataset includes over 3.5 million images and 3,749 lesions spanning more than 30 pathological subtypes.
This pretraining allows BUSGen to learn rich information of biological structures, pathological features, and clinical variations.
With few-shot adaptation, BUSGen could generate realistic and diverse task-specific data to facilitate the development of models for various downstream tasks, as demonstrated in \Cref{fig:framework}b.
Importantly, we make BUSGen publicly available to accelerate research progress in breast ultrasound analysis.

We demonstrate the \textbf{high adaptivity} of BUSGen for six essential downstream tasks associated with breast cancer screening, diagnosis and prognosis.
Trained on the generated data, the downstream models (dubbed ``BUS-DM" for BUSGen-based downstream model) consistently outperformed real-data-trained CLIP-based foundational models~\cite{radford2021learning} (dubbed ``Baseline-CLIP") across all downstream tasks.
For breast cancer screening, BUS-DMs achieved superior precision in detecting small nodules and classifying opportunistic screening-detected lesions compared with Baseline-CLIP.
For breast cancer early diagnosis, BUS-DM significantly outperformed board-certified radiologists (n=9), with an average sensitivity improvement of 16.5\% (P-value$<$0.0001) at the same specificity. Notably, it is the first time diagnostic models trained solely on generated data significantly outperformed experienced clinicians. Finally, BUS-DM outperformed Baseline-CLIP in predicting prognostic indicators, such as molecular subtype and axillary lymph node metastasis.

Moreover, our experiments highlighted several key properties of BUSGen.
First, it makes progress in patient \textbf{privacy protection}. In data generation, we leveraged recent advances to prevent generating images that are identical to real patient images in the training set, paving the way to completely de-privacy data sharing.
Additionally, we characterized a \textbf{scaling effect} of using generated data~\cite{fan2024scaling}, where BUSGen generated data was as effective as collected real-world data for training diagnostic models, evidenced by comparable performance scaling curves.
Finally, BUSGen improves the \textbf{generalization ability} of downstream models. We show that generated data prevents the spurious correlations stemming from the biases in data acquisition~\cite{ktena2024generative,ong2024shortcut}. As a result, BUS-DMs exhibit significantly smaller performance declines compared to their real-data-trained counterparts in external evaluations. 
In summary, BUSGen represents a transformative step in breast ultrasound analysis, providing a robust foundation for diverse applications while effectively addressing key challenges such as data scarcity, privacy concerns, and limited generalization ability. Our findings highlight its potential to revolutionize breast cancer screening, diagnosis, and prognosis.

\section{Results}
\label{sec:result}

\subsection{The BUSGen pretraining and adaptation framework}
\label{subsec:framework}

\begin{figure}[t]
    \centering
    \includegraphics[width=\linewidth]{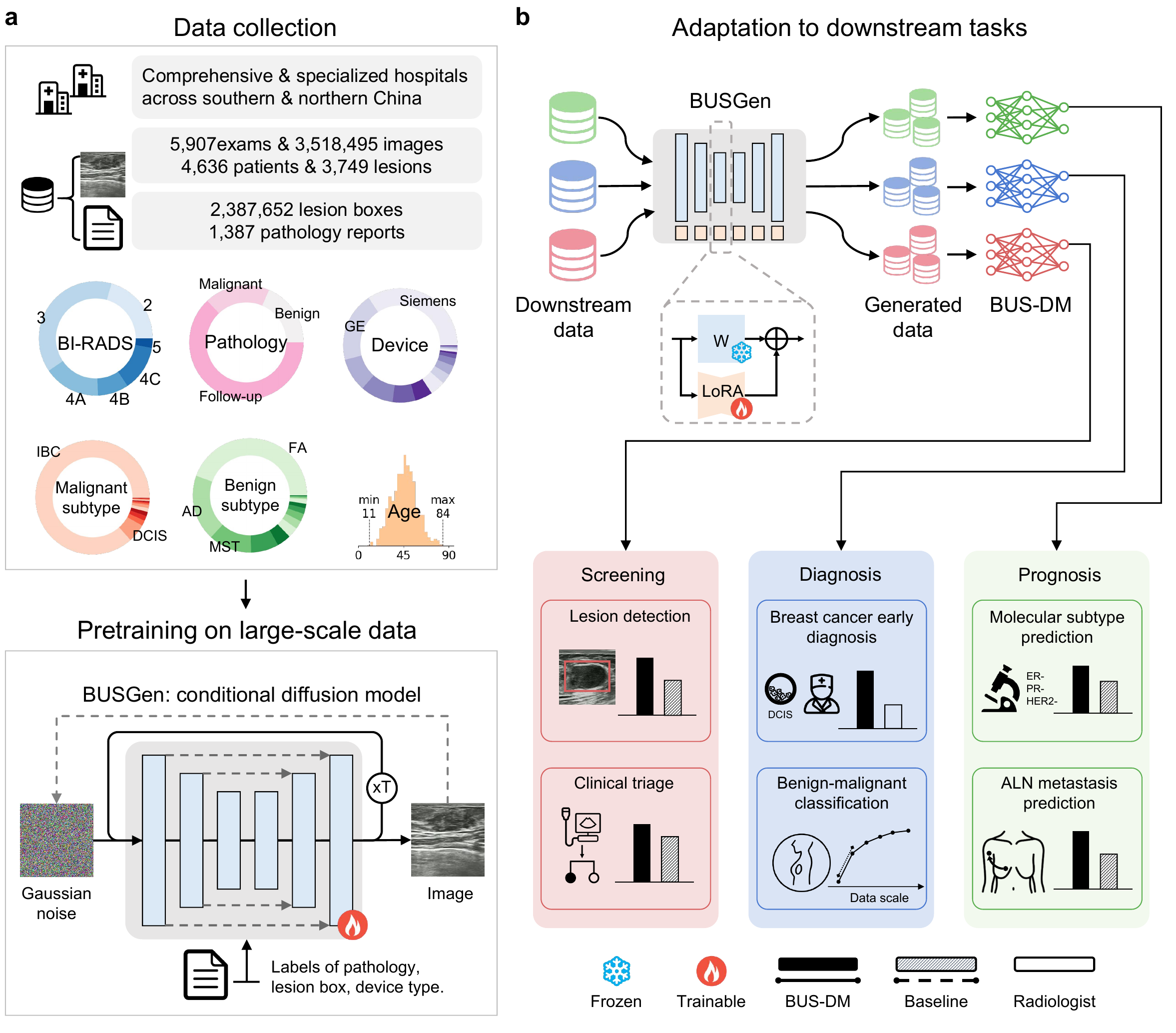}
    \caption{\small \small {\bf The schematic overview of the BUSGen pretraining and adaptation framework.} {\bf a}, Over 3.5 million breast ultrasound images of 5,907 examinations of 4,636 patients and 3,749 lesions were collected.
    These data were annotated by clinical experts and were used for the conditional generation task to pretrain the BUSGen model, enabling it to learn rich data distribution and generate high-quality images through an iterative refinement process repeated $T$ times.
    The pretraining task incorporated conditions of the labels of pathology, lesion box, and device type.
    {\bf b}, The pretrained BUSGen can be adapted to various downstream tasks, generating unlimited, informative data resources and facilitating the development of downstream models. To preserve the rich information acquired during pretraining, we froze the pretrained parameters and fine-tuned low-rank adapters (LoRA). In comparison to baseline models, the BUSGen-based downstream models (BUS-DMs) achieved superior performance in a wide range of tasks across breast cancer screening, diagnosis and prognosis.
    }
    \label{fig:framework}
\end{figure}

We developed BUSGen, a foundational generative model for breast ultrasound analysis, pretrained on extensive datasets and designed to generate unlimited, informative, and task-specific data to enhance the performance of deep learning approaches.
As shown in \Cref{fig:framework}, the ability of this foundational generative model was empowered by two key parts in our frameworks: pretraining and adaptation.

\textbf{Pretraining data.} We collected an extensive dataset from multiple institutions, including comprehensive and specialized cancer centers across various regions. The dataset integrates data from a diverse and representative patient population spanning different ages (11-84 years) and tumor subtypes, which were acquired by various types of scanner devices. As shown in \Cref{fig:framework}a, the constructed dataset (referred to as ``BUS-3.5M") contains 3,518,495 breast ultrasound images (1,130,843 normal images and 2,387,652 images with lesions) and 3,749 breast lesions. Details of BUS-3.5M are illustrated in Methods~\Cref{subsec:pretrain_data} and Supplementary Information Section 1. This extensive dataset provided a rich source of biological and clinical information for pretraining. Expert annotations were extracted from clinical reports and verified by clinical experts, offering ``gold standard" supervisory data, including lesion pathology (benign and malignant), lesion location (bounding boxes), and scanner device type. These detailed annotations provided valuable supervisory information that guided the pretraining process.

\textbf{Pretraining algorithms.} We designed the pretraining task to encourage the BUSGen model to capture the structures and textures of breast anatomy (e.g., skin, fat and glandular tissue), subtle pathological changes, and variations in clinical practices.
We designed BUSGen as a diffusion model~\cite{ho2020denoising,dhariwal2021diffusion}, capable of effectively capturing the distribution of large-scale data. 
We adopted a lightweight U-Net architecture (approximately 50M parameters), enabling fast speed and reduced GPU memory requirements, thus broadening its usability.
We employed the conditional generation task with classifier-free guidance algorithm~\cite{ho2021classifierfree} to encode valuable information as conditional embeddings within BUSGen. 
To capture high-frequency details of subtle structures, we pretrained BUSGen directly in pixel space~\cite{dhariwal2021diffusion}, rather than in low-dimensional latent space~\cite{rombach2022high,bluethgen2024vision}.
We conducted an ablation experiment to show the advantage of our model architecture and pretraining task design (Supplementary Information Section 2). Details of the pretraining algorithms are illustrated in Methods~\Cref{subsec:method_pretraining}.

\textbf{Adaptation algorithms.} After pretraining, we adapted BUSGen to multiple downstream tasks, as illustrated in \Cref{fig:framework}b. We developed several algorithms to enhance the fidelity and diversity of the generated task-specific data, particularly in few-shot adaptation settings.
To preserve the rich information acquired during pretraining, we froze the pretrained parameters of BUSGen and fine-tuned additional low-rank adapters (LoRA)~\cite{hu2022lora}. 
To prevent over-fitting, we introduced a device augmentation technique in addition to strong conventional data augmentations. We employed the DPMSolver++~\cite{lu2022dpm,lu2022dpm++} algorithm to accelerate BUSGen's sampling process while maintaining high image quality, and we used CPSampling~\cite{kazdan2024cpsample} to protect patient privacy.
The sampling process of BUSGen is efficient and scalable, which costs $\sim$2.14 seconds to generate one image using one RTX 4090 GPU.
We sampled abundant category-balanced downstream data and conducted data cleaning to remove low-quality samples (Methods~\Cref{subsec:method_adaptation} and Supplementary Information Section 3).
These high-quality generated data were then used to train models for downstream tasks. These models are referred to as BUS-DMs (BUSGen-based downstream models) in the following sections. Details of downstream methods are illustrated in Methods~\Cref{subsec:busgen_dm}.

To evaluate the efficacy of BUSGen, we compared BUS-DMs with real-data-trained CLIP-based foundational models~\cite{radford2021learning} (referred to as ``Baseline-CLIP"), where we leverage the model parameters pretrained on 1.65 million image-caption pairs extracted from medical papers at PubMed Central Open Access~\cite{lin2023pmc}.
We transferred them to the breast ultrasound domain by training on image-report pairs of the BUS-3.5M dataset and adapted to downstream tasks through fine-tuning on real downstream datasets. For a fair comparison, all baseline models shared the same vision encoder (ViT-L~\cite{dosovitskiy2021thomas}) as BUS-DMs. More details of Baseline-CLIP are illustrated in Supplementary Information Section 4.

\subsection{BUSGen generates realistic data while protecting privacy}

\begin{figure}[t]
    \centering
    \includegraphics[width=\linewidth]{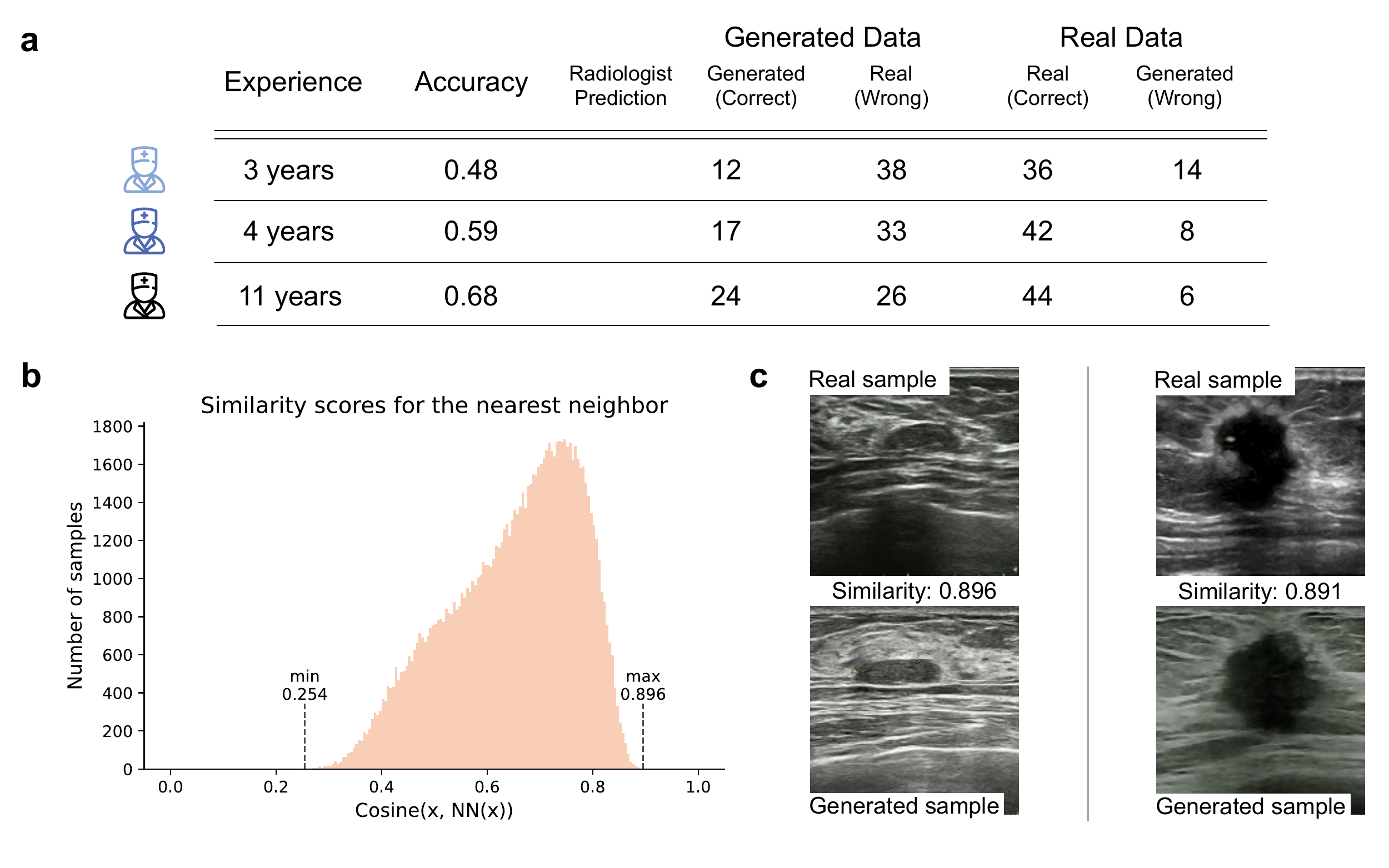}
    \caption{\small \small {\bf BUSGen can generate realistic data while protecting patient privacy.} 
    {\bf a}, Results of Visual Turing Test. Three radiologists were asked to distinguish ``fake" images generated by BUSGen from real images. They were presented with a set of 100 images, consisting of 50 generated and 50 real images. The results show that approximately 50\%-75\% of the generated images were mistakenly identified as real images by the radiologists.
    {\bf b}, Distribution of cosine similarity scores in the feature space between generated samples and their nearest neighbor in the training data. This result indicates that BUSGen will not replicate its training data, as the highest cosine similarity is 0.896.
    {\bf c}, Visualization of two pairs of generated and real images with the highest cosine similarity scores. As shown in the plots, these image pairs are not exact replications.
    }
    \label{fig:turing_test}
\end{figure}

To verify the fidelity of generated data, we conducted a Visual Turing Test~\cite{geman2015visual}, involving board-certified radiologists (n=3) with clinical experience of 3, 4, and 11 years, respectively. We presented 100 images, consisting of 50 generated and 50 real images and asked readers to distinguish ``fake" images generated by BUSGen from real images.
\Cref{fig:turing_test}a shows the predictions and accuracy of each reader's predictions as well as their corresponding experience. These results show that approximately 50\%-75\% of the generated images were mistakenly identified as real by the radiologists, indicating the biological structures and semantic features of more than half of the generated images were realistic for radiologists.

As BUSGen provides breast ultrasound images with high fidelity, a reasonable concern is patient privacy leakage when releasing this foundational generative model because diffusion models tend to exactly replicate images in their training sets~\cite{carlini2021extracting}.
We attempted to make progress in patient privacy protection and took advantage of the recent advances in sampling strategy (CPSampling~\cite{kazdan2024cpsample}) to protect the privacy of patients in our pretraining data.
To illustrate the effectiveness of privacy protection, we visualize the similarity scores of BUSGen-generated samples and their nearest neighbor in the training data in \Cref{fig:turing_test}b.
In addition, we present two image pairs with the highest similarity scores in \Cref{fig:turing_test}c. These results demonstrate that BUSGen will not exactly replicate any image in the training data, paving a new way to completely de-privacy data sharing.

\begin{figure}[t]
    \centering
    \includegraphics[width=\linewidth]{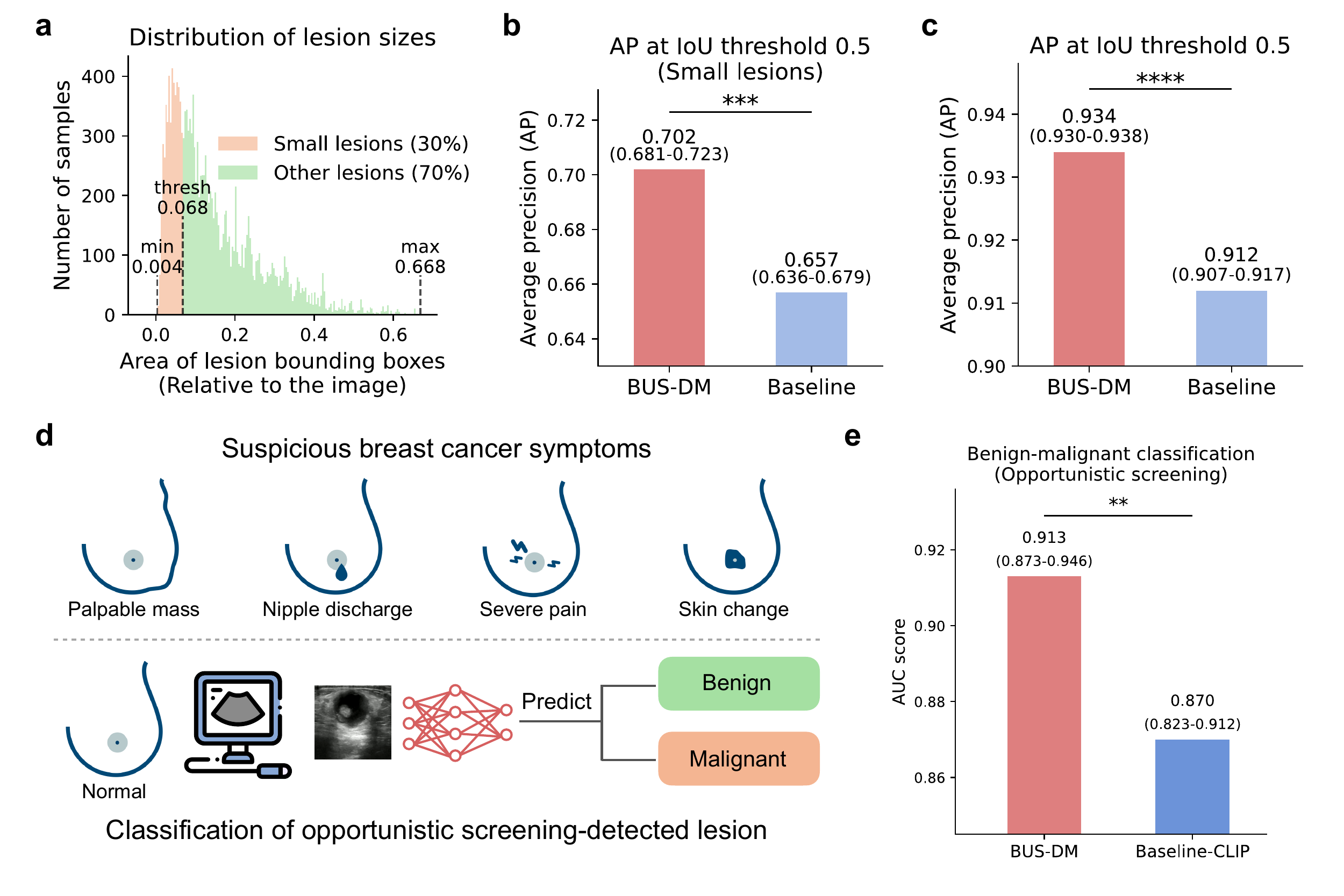}
    \caption{\small \small {\bf BUSGen improves breast cancer screening tasks.} 
    {\bf a}, Distribution of lesion sizes. The smallest 30\% lesions (orange) are defined as lesions with areas smaller than 0.068 (relative to images), which can be hard for radiologists to detect during ultrasound scanning.
    {\bf b}, Comparison of BUS-DM (red) and Baseline (light blue) in small lesion detection (n=16,896). We report Average Precision (AP) at an Intersection over Union (IoU) threshold of 0.5. BUS-DM achieved an AP$_{\text{small}}$ of 0.702 (95\% CI 0.681$-$0.720) and outperformed Baseline (blue; AP$_{\text{small}}$: 0.657; 95\% 0.637$-$0.679; P-value=0.0017).
    {\bf c}, Comparison of BUS-DM (red) and Baseline (light blue) in lesion detection (n=28,150). BUS-DM achieved an AP of 0.934 (95\% CI 0.930$-$0.938) and significantly outperformed Baseline (blue; AP: 0.912; 95\% 0.907$-$0.917; P-value$<$0.0001).
    {\bf d}, Illustration of opportunistic screening and classification tasks. The opportunistic screening is performed on a population without suspicious symptoms of breast cancer. The suspicious breast cancer symptoms include palpable mass, nipple discharge, severe pain and skin change (upper). Using deep learning models, we predict whether opportunistic screening-detected lesions are benign or malignant (lower).
    {\bf e}, Comparison of BUS-DM (red) with Baseline-CLIP (blue) for benign-malignant classification of opportunistic screening-detected lesions. BUS-DM achieved a higher AUC of 0.913 (95\% CI: 0.874–0.948) compared to Baseline-CLIP with an AUC of 0.870 (95\% CI: 0.823–0.912; P-value=0.0074). **P-value$<$ 0.01; ***P-value$<$ 0.001; ****P-value$<$0.0001.
    }
    \label{fig:screening}
\end{figure}

\subsection{Improving breast cancer screening}
\label{subsec:screening}

Breast cancer screening aims to detect breast cancer in a normal population without severe symptoms~\cite{elmore2005screening,myers2015benefits} where ultrasound is widely used for breast cancer screening~\cite{berg2012detection}, because of its high sensitivity to breast nodules, especially in the dense breasts~\cite{shen2015multi,brem2015screening,sood2019ultrasound,park2020pan}.
However, precise detection of small lesions and accurately interpreting screening-detected lesions are challenging in clinical practice~\cite{yuan2020supplemental}.
We adapted BUSGen to these two screening tasks to evaluate its efficacy in breast cancer screening.

The lesion detection task~\cite{yap2020breast,yap2017automated} aims to precisely identify whether the breast ultrasound images contain lesions and locate all lesions using bounding boxes.
As shown in \Cref{fig:screening}a, we evaluated the detection performance on small lesions which are defined as lesions with smallest 30\% relative areas (n=16,896; 4,822 with lesions and 12,074 without lesions). As shown in \Cref{fig:screening}b, for small lesion detection, our BUS-DM achieved an Average Precision (AP$_{\text{small}}$) of 0.702 (95\% CI 0.681$-$0.720), while the baseline model achieved an AP$_{\text{small}}$ of 0.657 (95\% CI 0.637$-$0.679; P-value=0.0017).
As shown in \Cref{fig:screening}c, on the whole test set (n=28,150; 16,076 with lesions and 12,074 without lesions), our BUS-DM achieved an AP of 0.934 (95\% CI 0.931$-$0.938); while the baseline model achieved an AP of 0.912 (95\% CI 0.907$-$0.917; P-value$<$0.0001).

Furthermore, to facilitate accurate clinical triage, we conducted experiments in interpreting opportunistic screening-detected lesions. These lesions were detected from patients who had no symptoms of breast cancer, such as palpable breast masses, nipple discharge, severe breast pain, or changes in breast appearance, as illustrated in \Cref{fig:screening}d.
In this task, our BUS-DM achieved an AUC of 0.913 (95\% CI 0.873$-$0.946) which significantly exceeded Baseline-CLIP (AUC: 0.870; 95\% CI 0.823$-$0.912; P-value=0.0074), as shown in \Cref{fig:screening}e. These results show the high adaptivity of BUSGen in breast cancer screening.

\subsection{Facilitating generalizable model for breast cancer early diagnosis}
\label{subsec:early_diagnosis}

Early diagnosis of breast cancer can significantly reduce mortality. Here, we conducted an experiment to distinguish an important early-stage breast cancer (ductal carcinoma in situ) from benign lesions. Ductal carcinoma in situ (DCIS) is a pathological subtype of non-invasive early-stage breast cancer where all cancer cells are confined within the basement membrane~\cite{pinder2010ductal,ernster1997increases,winchester2000diagnosis}.
DCIS lacks typical malignant features of invasive cancer and sometimes exhibits non-mass lesions or nodules with regular shape or circumscribed margins, as shown in \Cref{fig:diagnosis}c.
In clinical practice, distinguishing DCIS from benign lesions using ultrasound imaging has long been considered difficult for radiologists~\cite{watanabe2017ultrasound,kim2008correlation}.
Here, we adapted BUSGen in a few-shot setting (trained on only 34 DCIS examples) and evaluated the performance on an external test set.
As shown in \Cref{fig:diagnosis}d, our BUS-DM achieved an AUC of 0.900 (95\% CI 0.860$-$0.939) and significantly outperformed the Baseline-CLIP (AUC: 0.846; 95\% CI 0.787$-$0.902; P-value=0.0002).

\begin{figure}[t]
    \centering
    \includegraphics[width=\linewidth]{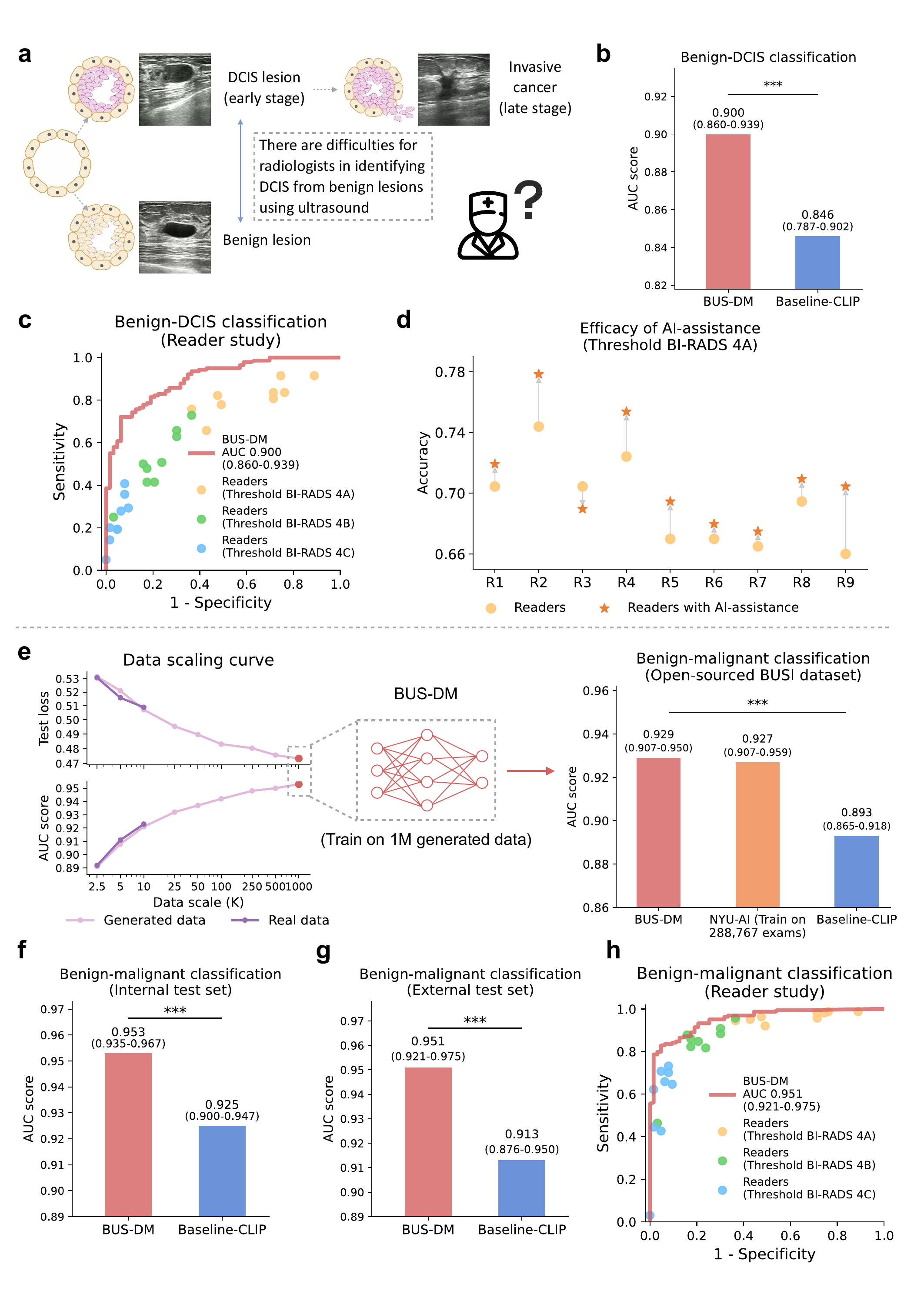}
\end{figure}
\begin{figure}[t]
    \centering
    \caption{\small \small 
    {\bf BUSGen enhances breast ultrasound diagnosis.}
    {\bf a}, Breast cancer early diagnosis involved the identification of DCIS (early-stage cancer) from benign lesions, which was considered difficult for radiologists based on ultrasound images.
    {\bf b}, Comparison of BUS-DM (red) with Baseline-CLIP (blue) in the early diagnosis task for benign-DCIS classification. BUS-DM achieved a higher AUC of 0.900 (95\% CI: 0.860–0.939) compared to the Baseline-CLIP with an AUC of 0.846 (95\% CI: 0.787–0.902; P-value=0.0002).
    {\bf c}, Comparison of BUS-DM with board-certified radiologists (n=9; 11 years of experience on average) in breast cancer early diagnosis. The ROC curves of BUS-DM (red curve) and diagnostic results of radiologists (dots) show that BUS-DM outperformed radiologists by a large margin.
    The colors (blue, green, and orange) of the dots represent radiologists' results calculated via different thresholds.
    {\bf d}, Accuracy improvements of radiologists with the assistance of BUS-DM. We report the accuracy of radiologists in breast cancer early diagnosis, as well as their accuracy after considering BUS-DM predictions. Accuracy is calculated using the threshold of BI-RADS 4A.
    {\bf e}, The data scaling curves of test loss (upper part of the left plot) and AUC score (lower part of the left plot) of diagnostic models trained on different scales of real collected data (dark purple) and BUSGen generated data (light purple). The curves for real and generated data closely align at small data scales, with the generated data continuously enhancing downstream performance as the number of generated samples increases.
    By scaling up the generated data to 1 million samples, we developed BUS-DM (AUC: 0.929; 95\% CI 0.907-0.950) that achieved comparable performance to NYU-AI (trained on 288,767 real samples; AUC: 0.927; 95\% CI 0.907-0.959), and outperformed Baseline-CLIP (AUC: 0.876; 95\% CI 0.849$-$0.903; P-value=0.0006) on the BUSI test set (n=780).
    {\bf f}, Comparison of BUS-DM (red) with Baseline-CLIP (blue) on the internal diagnosis test set for benign-malignant classification (n=579). BUS-DM achieved a higher AUC of 0.953 (95\% CI: 0.935–0.967) compared to the Baseline-CLIP with an AUC of 0.925 (95\% CI: 0.900–0.947; P-value=0.0006).
    {\bf g}, Comparison of BUS-DM (red) with Baseline-CLIP (blue) on the external diagnosis test set for benign-malignant classification (n=227). BUS-DM achieved a higher AUC of 0.951 (95\% CI: 0.921–0.975) compared to the Baseline-CLIP with an AUC of 0.913 (95\% CI: 0.876–0.950; P-value=0.0007). Note that BUS-DM, trained only on generated data, enjoyed better generalization ability than baseline models trained on real data.
    {\bf h}, Comparison of BUS-DM with board-certified radiologists (n=9) of the diagnosis task (benign-malignant classification) on the external test set. The ROC curves of BUS-DM (red curve) and diagnostic results of radiologists (dots) show that BUS-DM outperformed the average performance of radiologists.
    ***P-value$<$0.001.
    }
    \label{fig:diagnosis}
\end{figure}

To evaluate the practicality of BUS-DM, we compared it with nine board-certified radiologists with a range of experience of 3$-$26 years (11 years on average). These readers provided BI-RADS scores of the lesions, and we calculated their sensitivity and specificity using BI-RADS 4A as a threshold (details of the reader study are illustrated in Supplementary Information Section 5).
The results showed that our BUS-DM outperformed the average performance of readers by 16.5\% (95\% CI 11.7$-$21.5\%; P-value$<$0.0001) in sensitivity with the same specificity of 37.9\%; and by {43.0\% (95\% CI 32.8$-$53.6\%; P-value$<$0.0001)} in specificity with the same sensitivity of 81.3\%.
The performance of nine readers and BUS-DM are shown in \Cref{fig:diagnosis}e. After readers submitted their predictions, we provided them with BUS-DM results as an assistance and we showcased the changes of accuracy in \Cref{fig:diagnosis}f. Eight readers out of nine achieved better performance with the assistance of BUSGen, demonstrating the potential of our approach to improving the practicality of ultrasound imaging in breast cancer screening.

Note that BUS-DM, trained only on generated data, enjoyed better generalization ability than baseline models trained on real data.
Diverse and balanced device types of generated data may facilitate generation across clinical scenarios.
Additionally, we assumed that the generated data reduced the possibility of models learning spurious correlations derived from data acquisition biases (DAB)~\cite{geirhos2020shortcut,ong2024shortcut}. 
Following the previous work~\cite{ong2024shortcut}, we quantified the degree of DAB-induced shortcut learning and verified our assumption (Supplementary Information Section 6).

\subsection{Enhancing diagnosis with scalable generated data}
\label{subsec:diagnosis}

Diagnosis is an essential task in breast ultrasound analysis, requiring to classify the benignity or malignancy of lesions. As pathological features were already learned by BUSGen during the pretraining process, we directly sampled data without adaptation for training diagnostic models. First, we explored the critical question: when scaling up the generated data, to what extent could data generation replace data collection?
In \Cref{fig:diagnosis}e, we showcased the scaling effect~\cite{ahsan2021effect} of real and generated data where
we trained diagnostic models on various data scales and then evaluated their test losses and AUC scores. Restricted by computational resources, we scaled up generated data to 1 million samples. We highlight that our generated data were as effective as real collected data and continuously improved the diagnostic performance when scaling up, which could not be achieved in previous works~\cite{frid2018gan,sagers2023augmenting,fan2024scaling}.
We compared BUS-DM with an in-house NYU-AI which was developed with 288,767 real collected data.
Trained on 1 million generated images, our BUS-DM achieved comparable results (AUC: 0.929; 95\% CI 0.907$-$0.950) to the NYU-AI (AUC: 0.927; 95\% CI: 0.907$-$0.959) on an open-source BUSI test set, and significantly outperformed the Baseline-CLIP (AUC: 0.893; 95\% CI: 0.865$-$0.918; P-value=0.0006), as illustrated in \Cref{fig:diagnosis}e.

Further, we evaluated the efficacy of our approach on in-house test sets. In the internal evaluation (\Cref{fig:diagnosis}f), our BUS-DM achieved an AUC of 0.953 (95\% CI 0.935$-$0.967) which outperformed the Baseline-CLIP (AUC: 0.925; 95\% CI 0.900$-$0.947; P-value=0.0006).
Additionally, on a prospective external test set (\Cref{fig:diagnosis}g), BUS-DM achieved an AUC of 0.951 (95\% CI 0.921$-$0.975) which significantly outperformed the Baseline-CLIP (AUC: 0.913; 95\% CI 0.876$-$0.950; P-value=0.0007).
As shown in \Cref{fig:diagnosis}h, we compared BUS-DM with nine radiologists where BUS-DM outperformed the average reader performance by {30.3\% (95\% CI 19.2$-$41.1\%, P-value$<$0.0001)} in specificity at the same sensitivity of 96.4\% (at threshold BI-RADS 4A).

Further, we evaluated model performance on hard cases for radiologists, i.e., BI-RADS 4 lesions, which are suspicious malignant and recommended biopsies. Up to 75\% of breast biopsies in the United States finally resulted in benign findings~\cite{kerlikowske2003evaluation}. Therefore, patients would benefit from AI tools that could precisely interpret the BI-RADS 4 lesions. We evaluated BUS-DM on the BI-RADS 4 subgroup, which significantly reduced false positives.
Additionally, we assessed the performance gains across pathological subtypes and demonstrated that BUSGen consistently improved the baseline model performance across various benign subtypes. Details of subgroup analysis are illustrated in Supplementary Information Section 7.

\subsection{Facilitating prognostic indicator prediction}
\label{subsec:prognosis}

\begin{figure}[t]
    \centering
    \includegraphics[width=\linewidth]{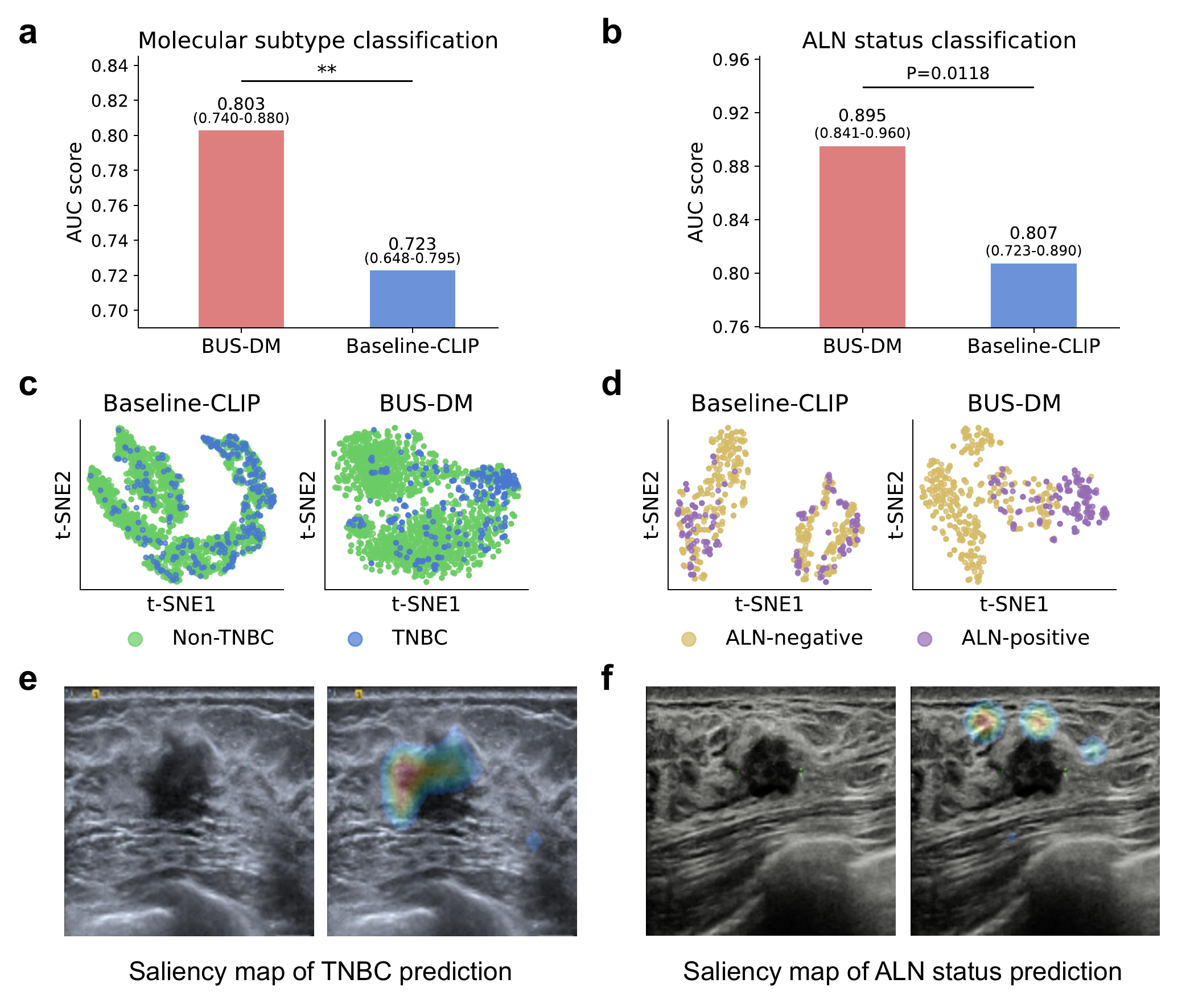}
    \caption{\small \small {\bf BUSGen facilitates breast cancer prognosis.} 
    {\bf a}, Comparison of BUS-DM (red) with Baseline-CLIP (blue) in molecular subtype classification (TNBC vs. non-TNBC). BUS-DM achieved a higher AUC of 0.954 (95\% CI: 0.932–0.983) compared to the Baseline-CLIP with an AUC of 0.723 (95\% CI 0.648$-$0.795; P-value=0.0046).
    {\bf b}, Comparison of BUS-DM (red) with Baseline-CLIP (blue) in ALN metastasis status classification (ALN-negative vs. ALN-positive). BUS-DM achieved a higher AUC of 0.895 (95\% CI: 0.841–0.960) compared to the Baseline-CLIP with an AUC of 0.807 (95\% CI 0.723$-$0.890; P-value=0.0118).
    {\bf c}, t-SNE plots of classification features (referred to as [CLS]) of downstream models in molecular subtype prediction. Clusters of TNBC and non-TNBC of BUS-DM [CLS] features are more concentrated than [CLS] features of Baseline-CLIP.
    {\bf d}, t-SNE plots of [CLS] features in the ALN status classification. Clusters of ALN-negative and ALN-positive BUS-DM [CLS] features are more concentrated than [CLS] features of Baseline-CLIP.
    {\bf e}, Saliency map of molecular subtype prediction by BUS-DM. The upper-left part of the lesion margin is highlighted by BUS-DM in predicting TNBC.
    {\bf f}, Saliency map of ALN status prediction by BUS-DM, which pays more attention to the surrounding glandular tissues of the lesion for predicting ALN metastasis. **P-value $<$ 0.01.}
    \label{fig:prognosis}
\end{figure}

Prognosis aims to predict the final outcomes of patients with breast cancer. It is difficult to directly predict the final outcomes~\cite{weigel2010current}, as many potential influencing factors are not well-defined or not available~\cite{baxevanis2021balance}. Instead, we conducted experiments to predict indicators that are highly related to prognosis. We adapted BUSGen to two prognostic indicator prediction tasks in the few-shot settings and evaluated the performance of BUS-DMs trained on datasets augmented by generated data.

We conducted experiments on predicting triple-negative breast cancer (TNBC), which is a molecular subtype of breast cancer that tends to be more aggressive and resistant to common treatments with a high recurrence rate and poor prognosis than non-TNBC~\cite{liedtke2008response,kennecke2010metastatic}.
The ``gold standard" molecular subtype labels are determined by immunohistochemistry (IHC) and in situ hybridization (ISH) confirmed biomarkers (ER-$\alpha$, PgR and HER2).
As shown in \Cref{fig:prognosis}a, the BUS-DM achieved an AUC of 0.803 (95\% CI 0.740$-$0.880) and outperformed the Baseline-CLIP (AUC: 0.723; 95\% CI 0.648$-$0.795; P-value=0.0046).

Additionally, we conducted experiments to predict axillary lymph node (ALN) status (positive or negative for ALN metastasis) which plays an essential role in treatment planning for breast cancer~\cite{rao2013axillary}, being the most significant prognostic indicator for early-stage patients~\cite{cianfrocca2004prognostic}. 
Pre-operative prediction of ALN metastasis from ultrasound images could pave the way for optimized clinical decision-making.
The ``gold standard" ALN status labels are determined by pathology after sentinel ALN biopsy or dissection.
As shown in \Cref{fig:prognosis}b, the BUS-DM achieved an AUC of 0.895 (95\% CI 0.841$-$0.960) while the Baseline-CLIP achieved an AUC of 0.807 (95\% CI 0.723$-$0.890; P-value=0.0118).

In \Cref{fig:prognosis}c and d, we provides the t-SNE plots of embeddings produced by different models. The plots show that BUS-DM distinguished TNBC from non-TNBC better than Baseline-CLIP, and BUS-DM separated ALN-positive from ALN-negative lesions more effectively than Baseline-CLIP.
For quantitative evaluation of clustering results, we employed metrics including normalized mutual information (NMI), adjusted Rand index (ARI) and silhouette coefficient (SIL) (Supplementary Information Section 8). The BUS-DM had higher NMI, ARI and SIL scores, showing its higher generalization ability than the Baseline-CLIP.

In an attempt to understand the decision-making process of BUS-DMs, we conducted a qualitative analysis of saliency maps. These saliency maps, obtained by GradCAM~\cite{selvaraju2017grad} indicated the important regions for BUS-DM to identify the prognostic indicators.
\Cref{fig:prognosis}e shows the saliency map of identifying a TNBC lesion. The upper left part of the lesion margin is highlighted by BUS-DM in predicting TNBC. 
In \Cref{fig:prognosis}f, we show the saliency map of predicting a lesion with positive ALN status for metastasis. Instead of focusing on the lesion structure, we found BUS-DM paid more attention to the surrounding glandular tissues of the lesion.
These intriguing findings might benefit in identifying biomarkers of ultrasound-based TNBC and ALN status prediction.

\section{Discussion}
\label{sec:discussion}

Data scarcity restricts the development of deep learning models in various breast ultrasound analysis tasks. 
At the same time, deep learning is undergoing a paradigm shift with the rise of foundation models~\cite{bommasani2021opportunities,bluethgen2024vision,wang2024self} trained on large-scale data that can be adapted to a wide range of downstream tasks.
This paradigm can be particularly advantageous for medical domains where access to high-quality annotated data is limited.
In this paper, we introduce BUSGen, the first foundational generative model in breast ultrasound image analysis, capable of learning features and patterns from over 3.5 million images and generating data tailored for various downstream tasks with the guidance of a few examples. These generated data can serve as unlimited resources to train downstream models.

We demonstrate the {high adaptivity} of BUSGen to improve a wide range of downstream tasks: lesion detection, opportunistic screening-detected lesion analysis, early-stage cancer diagnosis, lesion diagnosis, molecular subtype prediction, and metastasis prediction.
Moreover, for the first time, we characterized the {scaling effect} of generated data in the medical domain and show that generated data can be as effective as real collected data in diagnostic tasks.
Additionally, we find that BUSGen improves the {generalization ability} of downstream models, which is also reported in recent works~\cite{ktena2024generative}.
Finally, we make progress in patient {privacy protection} with generated data, and expect this de-privacy approach could encourage data sharing in more medical tasks.

Our study could potentially facilitate various downstream applications in real-world clinical practice.
First, our BUS-DM significantly outperforms human radiologists on DCIS, an important early-stage cancer subtype. This makes it suitable for integration into the workflow of breast ultrasound diagnosis to facilitate early identification of breast cancer.
Additionally, the BUS-DM for lesion diagnosis could re-evaluate a huge volume of retrospective breast ultrasound cases deposited in hospitals that have not undergone biopsies or surgeries, identifying potential false negatives and prompting further examinations. 
Finally, the BUS-DM for ALN metastasis prediction could be used before ALN biopsy or dissection, making a way for optimized clinical decision-making.
These improvements might contribute to better treatment outcomes and reduced mortality rates.

In the future, we will pretrain the BUSGen model on a more comprehensive dataset, covering breast ultrasound images under various drug treatment and post-operative conditions, to facilitate more tasks, such as prediction of the efficacy of neoadjuvant therapy response, recurrence rate and disease-free survival time.
Additionally, we will conduct experiments to explore more efficient and effective human-AI interactions of BUS-DMs in real-world clinical settings.
Finally, we expect the paradigm of foundational generative models to be extended to a wider range of clinical tasks across more imaging modalities, such as computed tomography and magnetic resonance imaging.

\section{Methods}
\label{sec:methods}

\subsection{Ethical approval}

Our study was approved by the institutional review board of the Peking University Cancer Hospital \& Institute (ID: 2024YJZ41). The study was not interventional and was performed under guidelines approved by the institutional review board. Informed consent was waived since the study presents no more than minimal risk. All datasets processed for this research were de-identified before being transferred to study investigators.

\subsection{Pretraining data collection, processing, and annotation}
\label{subsec:pretrain_data}

\subsubsection{Data collection}
\label{subsubsec:data_collection}
For pretraining the foundational generative model, we collected scanning videos of 7,965 breast ultrasound examinations of 5,985 patients from two institutions, Peking University Cancer Hospital \& Institute (PKUCH) and Nanchang People’s Hospital (NPH). These exams were conducted between January 2020 and March 2021. 
We randomly sampled an internal test set for lesion diagnosis (detailed in \Cref{methods:diagnosis}) and kept others in the pretraining set where we sampled by patients to prevent information leakage. This pretraining dataset covered a broad range of conditions in breast ultrasound analysis. The collected data covered patients aged 11 to 84, with lesions encompassing 34 pathological subtypes and 4 molecular subtypes.

\subsubsection{Data processing}

We removed low-quality data where clinical information is incomplete or lesions could not be clearly visualized.
After data processing, quality control and data sampling, we kept 4,636 patients (1,589 normal patients and 3,047 abnormal patients with 3,749 lesions). In total, the pretraining dataset contained 5,907 breast ultrasound examinations with scanning videos (2,157 normal videos and 3,750 videos with lesions) and 3,518,495 images (1,130,843 normal images and 2,387,652 images with lesions). Note that these massive images might contain redundant temporal information.

\subsubsection{Data annotation}

To incorporate more information in the pretraining process, we extracted rich ``gold standard" annotations by clinical experts from ultrasound reports. Based on pathological reports, 1,387 out of 3,749 lesions were attached with biopsy- or surgery-confirmed pathology labels (694 benign and 693 malignant), encompassing 1,454 videos (733 benign and 721 malignant) and 931,525 images (404,220 benign and 527,305 malignant). Additionally, clinical experts annotated bounding boxes to indicate lesion areas based on surgery or ultrasound reports, and they identified the device types of ultrasound scanners (9 manufacturers with 18 device types) from the ultrasound reports.

\subsection{BUSGen pretraining and adaptation framework}
\label{subsec:gen_model}

\subsubsection{BUSGen pretraining}
\label{subsec:method_pretraining}

We developed BUSGen as a conditional Denoising Diffusion Probabilistic Model (diffusion model)~\cite{ho2020denoising,dhariwal2021diffusion} with lightweight U-Net architecture.
The pretraining process of diffusion models enabled it to learn the data distribution $P(x)$ of breast ultrasound images. 
Specifically, diffusion models learn to gradually denoise a random Gaussian sample $x_T\sim \mathcal{N}(0, I)$ to a realistic image $x_0\sim P(x)$. This is achieved via learning the reverse process $P(x_{t-1}|x_{t})$ of a Markov Chain of length $T$. 
Diffusion models can be interpreted as a denoising autoencoder $\epsilon_{\theta}(x_t,t)$, which estimates the noise $\epsilon$ in $x_t$ at each step. The learning objective is simplified to:
\begin{align}
    \mathcal{L}=E_{x,\epsilon,t}\left[||\epsilon - \epsilon_{\theta}(x_t,t)||_2^2\right]
\end{align}
where $t$ is uniformly sampled from $\{1,\cdots,T\}$. 
BUSGen uses a conditional DDPM~\cite{ho2021classifierfree} $\epsilon_{\theta}(x_t,t,c)$ to learn the conditional data distribution $P(x|c)$ where $c$ can be any conditioning input. Here, we used the conditions of lesion boxes, pathology labels and device types where lesion boxes were encoded as positional embeddings, and pathology labels and device types were encoded as one-hot condition vectors.

\subsubsection{BUSGen adaptation}
\label{subsec:method_adaptation}

In adaptation of BUSGen to downstream tasks, we designed several algorithms to ensure the high quality of generated data, even in the case of limited downstream data.
To preserve the domain knowledge acquired during the pretraining process, we froze BUSGen parameters and fine-tuned only the additional lightweight parameters designed as low-rank adapters (LoRA). 
To prevent overfitting, we apply strong conventional data augmentations, such as random crop, color jittering, and random flip. 
Additionally, we incorporate a novel device type data augmentation, transforming each image to all device types~\cite{gatys2016image}. 
This image-to-image translation task was performed using CycleGAN~\cite{zhu2017unpaired} models, where each CycleGAN model was responsible for transferring images from any source device to the target device. For each real image, 18 augmented images from various device types were generated while preserving the same ``content".

We employed classifier-free guidance~\cite{ho2021classifierfree} to achieve better control of the generated images with input conditions. 
This algorithm could be formulated as:
\begin{align}
    \tilde{\epsilon}_{\theta}(x_t,t,c)=(1 + w)\epsilon_{\theta}(x_t,t,c)-w \epsilon_{\theta}(x_t,t)
\end{align} 
where $\tilde{\epsilon}_{\theta}(x_t,t,c)$ is defined as a weighted combination of conditional and unconditional outputs, and $w$ is a parameter that controls the strength of the guidance. 
The generation process of BUSGen is inherently slow due to the requirement of $T$ denoising steps. 
To accelerate the generation while maintaining high quality, we utilized DPM-Solver++~\cite{lu2022dpm,lu2022dpm++} as the sampling technique. 
Further, we designed a sampling strategy to ensure the diversity of generated data. 

As a probabilistic approach, BUSGen could occasionally generate low-quality data because the distribution $\hat{P}_{\theta}(x)$ learned from limited collected data was not perfectly aligned with the underlying data distribution $P(x)$. 
To address this issue, we designed a data-cleaning algorithm to remove the low-quality generated data. 
Specifically, we focus on generated images with incorrect labels (i.e., generated lesions that are inconsistent with the given conditions), as these can be particularly detrimental to the downstream model. 
We observed a notable improvement in the performance of the downstream model after the data-cleaning process.

\subsection{Downstream methods}
\label{subsec:busgen_dm}

The datasets used for the downstream tasks were collected from two external hospitals: Peking Union Medical College Hospital (PUMCH) and the Cancer Hospital of the Chinese Academy of Medical Sciences (CICAMS). However, an exception applies to the internal test set for the diagnostic task, which was collected from PKUCH and NPH (introduced in \Cref{subsubsec:data_collection}).
We conducted extensive experiments to verify the adaptivity of BUSGen. Specifically, we fine-tined BUSGen on various downstream tasks, and sampled generated images for training the BUS-DMs for various tasks. To select the hyperparameters, we adopted 5-fold cross-validation on downstream training sets and repeated each experiment 5 times for robustness.

\subsubsection{Breast cancer screening}
\label{subsubsection:methods_screening}
We conducted two tasks critical in breast cancer screening: (1) lesion detection and (2) clinical triage of opportunistic screening-detected lesions.
For the first task, the training set of lesion detection contained 67,911 images (38,425 with lesion boxes and 29,486 without lesions); and the test set contained 28,150 images (16,076 with lesion boxes and 12,074 without lesions). We directly produced generated images with lesion box conditions as BUSGen gained this ability during pretraining, and these box conditions were regarded as pseudo-labels in lesion detection. We used Mask R-CNN~\cite{he2017mask} models with ViT-L backbones as the detectors. We trained Baseline-CLIP on the real data and developed BUS-DM on the same dataset enhanced by 20,000 generated images.
For the second task, the opportunistic screening-detected lesions were collected from two external institutions. We divided the collected data evenly into training and test sets: the training set contained 265 lesions (154 benign and 111 malignant), and the test set contained 266 lesions (152 benign and 114 malignant). 
We adapted BUSGen to this task and generated 10,000 training data.
For this task, we developed the BUS-DM using only generated data. We first pretrained BUS-DM on 1 million generated training data mentioned in \Cref{subsec:diagnosis}, and then trained BUS-DM with these task-specific 10,000 generated data for the opportunistic screening task.

\subsubsection{Breast cancer early diagnosis}

We adapted BUSGen using 34 biopsy-confirmed DCIS lesions and trained the binary classifier using the generated DCIS and benign data. 
For this task, we developed the BUS-DM using only generated data with the same pipeline mentioned in \Cref{subsubsection:methods_screening}.
The test set contained 296 lesions (63 benign and 133 DCIS) where the DCIS lesions were purposely collected, and their corresponding examinations were conducted between January 2022 and April 2023. 
To compare BUS-DM with radiologists, and investigated how BUS-DM can assist radiologists, we conducted a reader study and invited nine board-certified breast ultrasound radiologists to analyze these lesions and provide their predicted BI-RADS scores. Note that we specifically informed readers that they should independently evaluate each lesion as the test data had different distribution with clinical practice. We calculated the sensitivity and specificity of readers using the BI-RADS 4A as the threshold for determining the binary predictions, i.e., we  assigned BI-RADS 2, 3 as benignity, and BI-RADS 4A$+$ as malignancy.

\subsubsection{Lesion diagnosis}
\label{methods:diagnosis}

For lesion diagnosis, the internal test set contained 579 lesions (274 benign and 305 malignant) with ``gold standard" labels. For external evaluation, we prospectively collected 227 lesions (63 benign and 164 malignant) from 225 consecutive patients who underwent breast ultrasound examinations and biopsies between October 2022 and March 2023. Additionally, we included a public Breast Ultrasound Images (BUSI) dataset~\cite{al2020dataset} collected from an institution in Egypt (437 benign, 210 malignant, and 133 negative lesions).
We trained Baseline-CLIP on the real data and developed BUS-DM with only generated data.
To evaluate the scaling effect, we generated 1 million images using BUSGen, and randomly sampled images in different data scales to train BUS-DMs where we repeatedly sampled 5 times at each data scale for robust evaluation.

\subsubsection{Prognostic indicator prediction}

For molecular subtype classification, the data were collected from two external institutions. The ``gold standard" labels were extracted by clinical experts based on IHC and ISH reports. We split all collected data into training and test sets. The training set contained 423 lesions (363 non-TNBC and 60 TNBC) and the test set contained 426 lesions (369 non-TNBC and 57 TNBC).
For ALN status classification, the data were also collected from two external institutions. The ``gold standard" labels were assigned by clinical experts based on ALN biopsy or dissection results, with negative ALN status defined as the absence of metastasis in any ALN. We split all collected data into training and test sets. The training set contained 110 lesions (77 ALN-negative and 33 ALN-positive) and the test set contained 117 lesions (80 ALN-negative and 37 ALN-positive). We trained Baseline-CLIP on the real data and developed BUS-DM with the training set enhanced by 10,000 samples generated by the BUSGen model, which had been adapted to these tasks, respectively.

\subsection{Statistical analysis}

We estimate the 95\% confidence intervals by 1,000 bootstrap replications. We calculate the two-sided P-values for significance comparisons of sensitivity and specificity using permutation tests with 10,000 permutations. The P-values of AUC are calculated using DeLong's test~\cite{delong1988comparing,sun2014fast}. 

\subsection{Implementation details}

We implemented the project based on the following packages: Python (3.9), OpenCV (4.9.0.80), Pandas (2.2.1), Numpy (1.26.4), and Pillow (10.3.0). Additionally, the deep learning model is implemented using PyTorch (1.10.1) and Torchvision (0.11.2). Evaluation metrics are calculated using Sklearn (1.4.1). We conducted the experiments using computational resources from seven GPU clusters. Four of these clusters were equipped with eight NVIDIA RTX 3090 GPUs each, while the remaining three featured eight NVIDIA RTX 4090 GPUs each.
The sampling process costs $\sim$2.14 seconds for generating one image using one RTX 4090 GPU. Therefore, the computational cost of generating 1 million images is $\sim$600 GPU hours.

\backmatter

\bmhead{Data Availability}

The BUSI dataset used in this study is publicly available at \href{https://www.kaggle.com/datasets/aryashah2k/breast-ultrasound-images-dataset}{https://www.kaggle.com/datasets/aryashah2k/breast-ultrasound-images-dataset}. We release a repository of the BUSI dataset with device augmentation.
However, due to respective Institutional Review Boards' restrictions and to protect patient privacy, the BUS-3.5M and the downstream datasets used in this study cannot be made publicly available. De-identified data may be made available by the corresponding authors for research purposes upon reasonable request.

\bmhead{Code Availability}

The pretraining and adaptation code for BUSGen and an online API will be released. Also, we provide an online demo of BUSGen at \href{https://aibus.bio/}{https://aibus.bio}.

\bmhead{Acknowledgments}
We thank Ruichen Li and Jigang Fan for their helpful suggestion and discussion.
Di He is supported by National Key R\&D Program of China (2022ZD0160300) and National Science
Foundation of China (NSFC62376007). Liwei Wang is supported by the National Science Foundation of China (NSFC62276005).

\bmhead{Author contributions}

H.Yu. and L.W. conceived and designed the study. 
Z.N., B.T., Y.Lu. and X.G. carried out data acquisition. 
Y.Li., H.Yu., Y.Lu., Z.N. and Q.W. carried out data processing and annotation.
H.Yu. developed the AI models.
Y.Li. carried out generated data cleaning.
Y.Li. developed the platform for reader study.
N.Z., Z.N., W.Q., J.T., M.Z., X.G., J.H., L.H. and Y.W. participated in the reader study.
H.Yu., H.Ye., S.H., D.H., Y.Li., N.Z., Z.N., D.W., Z.Z., Q.W., D.D., Q.Z., J.Z. and L.W. wrote and revised the paper.

\bmhead{Competing Interests}

The authors declare no competing interests.

\bibliography{sn-bibliography}

\end{document}